\documentclass[PHME, 2021]{PHMSociety}

\usepackage{graphicx}
\usepackage{amsmath}
\usepackage[utf8]{inputenc} 
\usepackage{amsmath}
\usepackage{amssymb}
\usepackage{mathrsfs}
\usepackage{graphicx}
\usepackage{xcolor}
\usepackage{float}
\usepackage{microtype}
\usepackage{amssymb} 
\usepackage{multicol}
\usepackage{fancyhdr} 
\usepackage{eqnarray}
\usepackage{subcaption}
\usepackage{lipsum}
\usepackage{titlesec}
\usepackage{pgfgantt}
\DeclareMathAlphabet\mathbfcal{OMS}{cmsy}{b}{n}
\usepackage{etoolbox}   
\usepackage{verbatim}   
\usepackage{color}
\newcommand{\tens}[1]{%
  \mathbin{\mathop{\otimes}\displaylimits_{1}}%
}
\definecolor{c1}{HTML}{808080} 
\definecolor{c2}{HTML}{8FED8F} 
\definecolor{c3}{HTML}{4DBA17} 
\definecolor{c4}{HTML}{ADD9E6} 
\definecolor{c5}{HTML}{FF9E00} 
\definecolor{c6}{HTML}{B3804D} 
\definecolor{c7}{HTML}{FF0080} 
\definecolor{c8}{HTML}{99994C} 
\usepackage{makecell}
\usepackage{multirow}
\usepackage{mathptmx}

\begin{document}

\title{Canonical Polyadic Decomposition and Deep Learning for Machine Fault Detection}

\author{%
	Frusque Gaëtan\authorNumber{1}, Michau Gabriel\authorNumber{2}, and Fink Olga\authorNumber{3}
}

\address{
	\affiliation{{1,2,3}}{ETH Zurich, Swiss Federal Institute of Technology, Zurich, Switzerland}{ 
		{\email{frusque.gaetan@outlook.fr}}\\ 
		} 
}

\maketitle
\pagestyle{fancy}
\thispagestyle{plain}

\phmLicenseFootnote{Gaëtan Frusque}

\begin{abstract}
Acoustic monitoring for machine fault detection is a recent and expanding research path that has already provided promising results for industries. However, it is impossible to collect enough data to learn all types of faults from a machine. Thus, new algorithms, trained using data from healthy conditions only, were developed to perform unsupervised anomaly detection. A key issue in the development of these algorithms is the noise in the signals, as it impacts the anomaly detection performance. In this work, we propose a powerful data-driven and quasi non-parametric denoising strategy for spectral data based on a tensor decomposition: the Non-negative Canonical Polyadic (CP) decomposition. This method is particularly adapted for machine emitting stationary sound. We demonstrate in a case study, the Malfunctioning Industrial Machine Investigation and Inspection (MIMII) baseline, how the use of our denoising strategy leads to a sensible improvement of the unsupervised anomaly detection. Such approaches are capable to make sound-based monitoring of industrial processes more reliable.
\end{abstract}

\section{Introduction}
\label{sec:intro} 

Early machine fault detection is a highly relevant industrial problem and is one of the pillars of the industry 4.0 to lower the cost of maintenance \cite{stojanovic}.  
Sound is one type of signals that can be recorded to monitor the health state of the machine. These signals are currently a topic of interest because acoustic monitoring offers a great benefit for the industries since it is cheap, non-invasive, non-disruptive and easy to install also as a refurbishment to existing assets.  
However, the discrimination between sounds from healthy and abnormal machines remain a challenging task due to following reasons: (1) failures are typically rare, (2) failures recorded are not necessarily representative of all  possible failure types, (3) the noise levels and the presence of artefacts can be high {\cite{purohit2019mimii}}, {\cite{chao2020fusing}}, (4) different operational modes make it difficult to characterize normal from abnormal states of the machine.

To tackle these problems, unsupervised anomaly detection strategies were proposed  \cite{purohit2019mimii}, \cite{michau2017deep}, \cite{tagawa2015structured}. The idea is to use healthy data only for the training of the anomaly detector. 
These approaches are designed to identify anomalies without any knowledge on the characteristics of the faulty system conditions. This method allows taking into account the imbalance between the two classes (healthy and abnormal case) as well as the lack of recordings of all possible faults. 

Deep learning based methods have recently shown remarkable performances for unsupervised detection problem (and more generally for audio signal processing  \cite{purwins_deep_2019}). These methods usually are a 1-class classifier \cite{michau_feature_2020} or an autoencoder with a decision on the residual to perform detection \cite{purohit2019mimii}. In \cite{michau_feature_2020}, a stacked extreme machine learning procedure is used to perform simultaneously unsupervised feature learning and feature aggregation in a single indicator for fault detection. This method can learn the relevant features from real condition monitoring data with higher accuracy compared to manually selected features. 
More recently \cite{michau2021unsupervised}, a 1-class classifier was combined with an adversarial domain discriminator to transfer the developed models and the operational experience between machines subject to changing operating conditions, in a setup where the target machine has only limited operational experience.

The condition monitoring signals are often recorded at a high sampling rate, and the number of samples is often too high for applying end-to-end deep learning methods on the raw signals directly. Feature extraction is then often applied for high frequency condition monitoring data. One possible way to extract informative features is to transform the signals in a different domain. Transformation in the frequency domain, via a spectrogram for example \cite{purohit2019mimii}, has been shown to be particularly suitable. However, more specific transformations using wavelet analysis or dictionary learning have been proposed \cite{whitaker_combining_2017}, \cite{chen_classification_2020}. In addition to reducing the dimension of the signal, a good signal transform highlights the relevant motifs characterising the dataset and facilitates the learning of the neural network. 

However, these transformations do not necessarily eliminate the noise and artefacts present in the dataset. A signal corrupted by several sources of noise will result in a poor anomaly detection performance by increasing the overfitting effect or by blocking the neural network training in irrelevant local minima. 

In this work, we focus on the importance of denoising the high frequency data before using a deep learning method for a subsequent anomaly detection task. The core idea of our proposed strategy is to reduce the complexity of the dataset, thereby, improving the classification performance of the neural network. To improve the quality of the spectral representation, we take advantage of the data structure, consisting of discontinuous recordings of the machine at several instants and focus on eliminating recordings-specific artefacts as well as noise. We propose a powerful, data-driven and quasi non-parametric denoising strategy using the non-negative Canonical Polyadic (nnCP) decomposition \cite{lim2009nonnegative}. This procedure is particularly suited when the sound of the machine is stationary. We compare this method to another denoising strategy using a more standard decomposition: the Non-Negative Matrix (NMF) factorisation \cite{lee_algorithms_2001}.

This article is organised as follows: we first introduce, in section \ref{S-met}, the proposed denoising strategies using the NMF and nnCP decomposition. These methods are then applied to a case study, the Malfunctioning Industrial Machine Investigation and Inspection (MIMII) baseline, which is presented in section \ref{S-cas}. Comparison and evaluation of the different methods using the MIMII dataset are presented in section \ref{S-res}. Finally, results and perspectives are discussed in section \ref{S-con}.

\section{Method}\label{S-met}
We consider a typical dataset used as the input feature of an 1-class classifier or an autoencoder for fault detection; which is the time-frequency representation of each audio recording from a machine. This dataset can be conveniently written as a three-dimensional cube of data, called tensor and denoted by $\mathbfcal{X}\in \mathbb{R}^{F \times T \times N} $, where $F$ is the number of frequencies considered, $T$ is the number of frames and $N$ is the number of recordings.  We present in this section two strategies using the NMF and the CP to denoise and remove artefacts in $\mathbfcal{X}$. 
\subsection{Non-Negative Matrix Factorisation (NMF)}\label{nmf}
NMF is a well-known matrix decomposition that can be seen as an instance of soft clustering \cite{aggarwal_data_2013}. It can be used for feature extraction and was recently applied on acoustic signals for event detection \cite{komatsu_acoustic_2016}. Since this method has to be applied on a matrix and not a tensor, we consider the matrix representation of the tensor $\mathbfcal{X}$, noted by $\mathbf{X}_{MAT} \in \mathbb{R}^{F \times TN}$ and defined as :
\begin{align}\label{E-mat}
\mathbf{X}_{MAT} = [\mathbf{X}_{::1} , \mathbf{X}_{::2}  , ..., \mathbf{X}_{::N}  ].
\end{align}
Notice that $\mathbf{X}_{MAT}$ is the concatenation of the time-frequency representations of each recording. Applying the NMF on $\mathbf{X}_{MAT} $ aims to find two matrices of lower dimensions $\mathbf{A} \in \mathbb{R}^{F \times K} $ and $\mathbf{B}\in \mathbb{R}^{TN \times K} $ which are the solution of the following optimisation problem :
\begin{equation}\label{CNMF}
\underset{\mathbf{A}>0, \mathbf{B}>0 }{\rm argmin} \hspace{0.1cm}   \mid\mid \mathbf{X}_{MAT}  - \mathbf{A} \mathbf{B}^t \mid\mid_F^2 
\end{equation} 
Where $\mid \mid \bullet \mid \mid_F$ is the Frobenius norm. $\mathbf{A}$ and $\mathbf{B}$ are constrained to be non-negative matrices (i.e. all their elements have to be higher or equal to zeros). The only hyperparameter here is $K$, the number of columns (or components) in each matrix. 

The columns of the matrix $\mathbf{A}$ can be seen as the main spectral motifs characterising the dataset. Each motif activates through time according to the corresponding activation profile which are the columns of the matrix $\mathbf{B}$. The non-negativity constraint is important here since it will induce sparsity in the obtained motifs which contribute to obtaining denoised components \cite{frusque_multiplex_2020}. Also, since spectral data are non-negative, it will help to obtain interpretable components \cite{donoho2004does},  \cite{frusque_semi-automatic_2020}. 

Thus, the idea of denoising with NMF, is to reconstruct the input feature $\mathbf{X}_{MAT}$ using the denoised motifs in the matrix $\mathbf{A}$. The denoised version of $\mathbf{X}_{MAT}$ using NMF is noted as $\hat{\mathbf{X}}_{MAT}$ and corresponds to :
\begin{equation}\label{ANMF}
\hat{ \mathbf{X}}_{MAT} = \mathbf{A} \mathbf{B}^t.
\end{equation} 
Please note that by performing the inverse transformation of Eq.~\ref{E-mat}, the denoised data from the matrix $\hat{ \mathbf{X}}_{MAT}$ can also be represented in the form of a tensor, noted as $\hat{ \mathbfcal{X}}$.

There are many algorithms to solve the optimisation problem stated in Eq.~\ref{CNMF}. The major difficulty here is that Eq.~\ref{CNMF} is non-convex. Multiplicative update \cite{lee_algorithms_2001} or alternating least square \cite{cichocki2007regularized} are state of the art approaches and are widely used algorithms to find a solution to the problem Eq.~\ref{CNMF}.

\subsection{Non-Negative Canonycal Polyadic Decomposition (nnCP)}
\begin{figure}[h]
\centering
\includegraphics[width=\columnwidth]{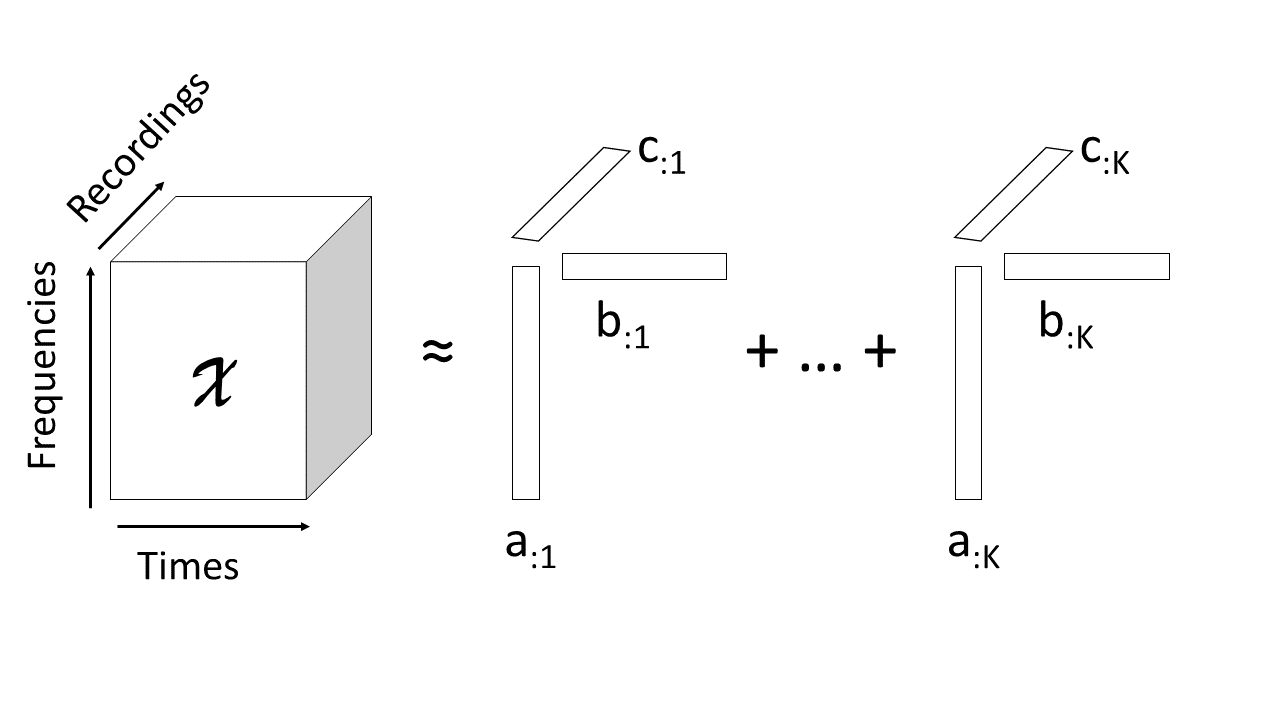}
\caption{Illustration of the canonycal polyadic decomposition.}\label{F-CP}
\end{figure}

We propose to take advantage of the characteristic 3-dimensional structure of the tensor $\mathbfcal{X}$. Tensor decompositions like canoncal polyadic decomposition are known to have good patern extraction properties. A canoncal polyadic decomposition was rencently used for fault detection in vibration data \cite{hu_tensor-based_2019}. The nnCP can then be seen as an extension of the NMF for data represented as a tensor. Before introducing the nnCP, it is important to note that the product between two matrices can be written as the sum of rank-1 matrices such as :
\begin{align}
 \mathbf{A} \mathbf{B}^t = \sum_{k=1}^{K} { \mathbf{a}_{:k} \times \mathbf{b}_{:k} }
\end{align}
With $\times$ being the vector product (let $\mathbf{u} \in \mathbb{R}^{F} $ and $\mathbf{v} \in \mathbb{R}^{T} $, the tensor product $\mathbf{u} \times \mathbf{v} =
\mathbf{M} \in \mathbb{R}^{L \times T}$ is such that $m_{ft} = u_{f}v_{t}$ \cite{comon2014tensors}).
Then, applying a nnCP decomposition on $\mathcal{X}$ aims to approximate this tensor with three matrices $\mathbf{A}\in \mathbb{R}^{F \times K} $, $\mathbf{B}\in \mathbb{R}^{T \times K} $ and $\mathbf{C}\in \mathbb{R}^{N \times K} $ maximising the following optimisation problem :
\begin{align}\label{CCP}
\underset{\mathbf{A} >0,\mathbf{B}>0, \mathbf{C}>0}{\rm argmin} \hspace{1cm}   \mid\mid  \mathbfcal{X}  - \sum_{k=1}^{K} { \mathbf{a}_{:k} \times \mathbf{b}_{:k} \times \mathbf{c}_{:k}     }  \mid\mid_F^2, 
\end{align}
The motivation behind the non-negativity constraint is the same as for the NMF. Moreover, it warrants the existence of a global solution for this decomposition \cite{qi2016uniqueness}. An illustration of the nnCP is provided in Figure~\ref{F-CP}.

This decomposition leads to remove several sources of noise and artefacts that are  specific to the particular recordings. Indeed, a component of the matrix $\mathbf{C}$, noted $\mathbf{c}_{:k} $, regroups recordings that share a similar spectral motif $\mathbf{a}_{:k}$. This spectral motif has to activate synchronously through time for every recording of the group (coinciding with the temporal activation $\mathbf{b}_{:k}$). Thus, only time-coherent motifs that are redundant in several recordings will be extracted by this decomposition. Moreover, the nnCP is particularly beneficial if the sound emitted by the machine is stationary. In that case, the spectral motifs characterising the signal will be associated with constant activation profiles present in several recordings. 

Finally, denoising with nnCP corresponds to reconstructing the input features $\mathbfcal{X}$ using the denoised motifs $\mathbf{A}$. The denoised version of $\mathbfcal{X}$ using nnCP is :
\begin{equation}\label{ACP}
\hat{ \mathbfcal{X}} =\sum_{k=1}^{K} { \mathbf{a}_{:k} \times \mathbf{b}_{:k} \times \mathbf{c}_{:k}}.
\end{equation} 

Several algorithms have been proposed to solve the optimisation problem Eq.~\ref{CCP} which is also non-convex. In this work, we used the Hierarchical Alternating Least Square (HALS) algorithm \cite{cichocki2009fast} as it is easy to implement and showed a good performance compared to other algorithms.

\section{Case study}\label{S-cas}
We apply our porposed denoising procedure to an acoustic monitoring case study, which is the benchmark from the MIMII dataset \cite{purohit2019mimii}. 
This dataset consists of four different machine operating sounds recorded in real factory environments. The four machines are a pump, a fan, a slider and a valve.
The benchmark consists of an autoencoder-based anomaly detector with  the log-Mel spectrogram of the sound signal as input features. In the following, we provide more details about this benchmark case study.
\subsection{Time-frequency representation}
The sound of each machine are recorded via an 8-channel microphone array. Only the signal from the first channel is considered. Each recording corresponds to a 10-second sound segment sampled at 16 kHz. To reduce the dimension of the input feature, a time-frequency representation of each recording is performed via a log-Mel spectrogram \cite{mcfee2015librosa}. Compared to a classic spectrogram, a Mel-spectrogram simulates the sound perception of humans. Indeed, humans do not perceive frequencies on a linear scale and are better at detecting differences in lower frequencies than higher ones. Then, in a Mel-spectrogram, the frequencies are rescaled (via the Mel scale) such that equal distances in pitch sounds similarly distant to the listener. This transformation is relevant if the difference between normal emission and anomaly can be easily perceived by a human with trained ears. 

The application of the log-Mel spectrogram on the $N$ recordings from one machine yields the following dataset denoted $\mathcal{X}\in \mathbb{R}^{F \times T \times N} $, with $F$ and $T$ respectively the number of Mel-frequencies and frames used for the log-Mel spectrogram computation of each sound recording. We use the same hyperparameters to compute the Mel-spectrogram as in \cite{purohit2019mimii}: we set $F=64$ and $T=314$. 
\subsection{Anomaly detector}\label{S-det}
The anomaly detector is based on an autoencoder. Considering the input feature $\mathbf{x} = \mathbf{x}_{:tn}$ $\forall t \in {1,...,T}$ and $\forall n \in {1,...,N}$ which is the log-Mel-frequencies of one frame from one recording, the loss function associated with the autoencoder is given by:
\begin{align}
 \mathcal{L}(\mathbf{x}; \theta)  = \mid \mid \mathbf{x} - D_{\theta_e}(E_{\theta_d}(\mathbf{x}))   \mid \mid_F^2,
\end{align}
where $\theta=(\theta_e,\theta_d)$, $E_{\theta_e}(\bullet)$ and $D_{\theta_d}(\bullet)$ are the encoder and the decoder networks, with parameters $\theta_e$ and $\theta_d$ respectively. The encoder network $E_{\theta_e}(\bullet)$ comprises two dense layers of 64 nodes plus one of 8 nodes. The decoder is composed of two dense layers of 64 nodes. The activation function used is the Rectified Linear Units (ReLUs).

In order to learn the distribution of the normal signals, the autoencoder is trained on sound from healthy machines only, referring to this training dataset as $\mathcal{X}^{train}$ . We are looking for the best parameters $\theta^{*}$ which minimise the root mean square error between input features and their reconstructions : 
\begin{align}\label{E-cost}
\underset{\theta}{\rm argmin} \hspace{0.2cm} \sum_{ \mathbf{x} \in \mathcal{X}^{train} }^{} {     \mathcal{L}(\mathbf{x}; \theta)    }  , 
\end{align}

The trained model will have a high reconstruction error when applied to an abnormal machine signal because it will be outside the distribution of the training samples. The separation between normal and abnormal sound can be made by thresholding the reconstruction error. More explicitly, considering a recording $\mathbf{X}$, we use the following decision function $\mathcal{F}(\mathbf{X},\theta^{*})$ to determine if the recording is from a normal or abnormal machine :
\begin{align}\label{E-des}
\mathcal{F}(\mathbf{X},\theta^{*}) = \begin{cases}  & \text{Normal if      } \sum_{ \mathbf{x} \in \mathbf{X} }^{} {   \mathcal{L}(\mathbf{x}; \theta^{*})  }   < \phi , \\
&  \\
 &   \text{Anormal if      } \sum_{ \mathbf{x} \in \mathbf{X} }^{} {   \mathcal{L}(\mathbf{x}; \theta^{*})  }   \geq \phi.
  \end{cases}
\end{align}
The selection of threshold $\phi$ is a compromise between detecting the anomalies and limiting the false detection.
This can be done by using a validation dataset containing abnormal machine sound or by analysing the distribution of the reconstruction error of the training dataset.

\subsection{Insertion of the denoising strategy}
\begin{figure}[H]
\centering
\includegraphics[width=\columnwidth]{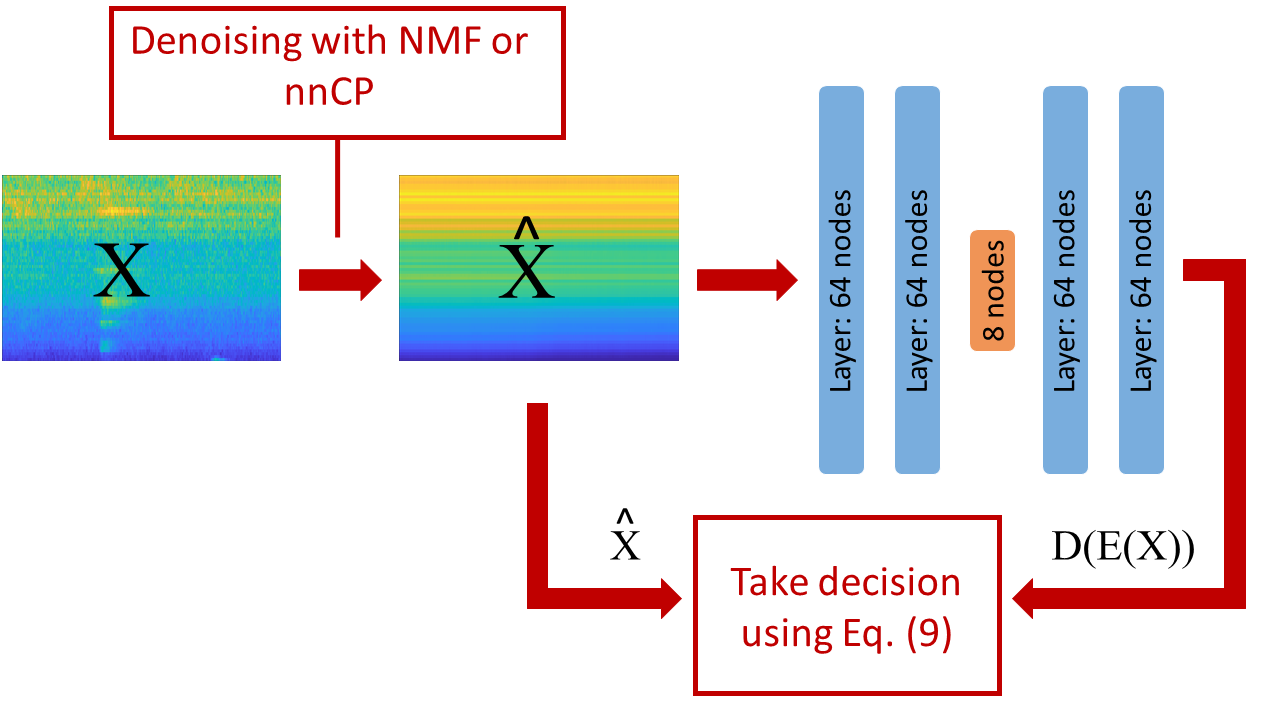}
\caption{Illustration of the detection method, the input feature is first denoised using the NMF or nnCP strategy, a decision is made by comparing the output of the autoencoder with the denoised input using Eq.~\ref{E-des}}\label{F-pre}
\end{figure}

To improve the autoencoder performance, we propose to use the denoised signal $\hat{ \mathbfcal{X}}$ derived by NMF or nnCP as input to the autoencoder. An overview of the proposed denoising strategy is provided Figure~\ref{F-pre}. We use the same autoencoder as in the baseline to make sure that the observed improvement is due to the effect of denoising applied on the input. The only hyperparameter to consider for each method is the number of components $K$. If $K$ is too low, the reconstruction is done with few spectral motifs. In this case, we will have good denoising properties and artefact removal but a crucial spectral motif can be missed. On the contrary, if $K$ is too high, the noise and artefacts present in the dataset will start to be reconstructed. However, several values of $K$ leads to relevant results. An Elbow criterion can be used to help with the selection of this hyperparameter.

Finally, two additional remarks have to be discussed: \\
$\bullet$ The log-Mel spectrogram contain only non-positive elements whereas NMF and nnCP have the non-negativity constraint. Then, the decomposition Eq.~\ref{CNMF} and Eq.~\ref{CCP} are performed on the absolute value of $\mathbfcal{X}$.\\ 
$\bullet$ The MIMII baseline used a data-augmentation strategy, where the input feature is not directly $\mathbf{x} $ (or $\hat{ \mathbf{x} }$), but the concatenation of five successive frames from the same recording. We use the same procedure here.


\section{Results}\label{S-res}
We consider the openly available dataset MIMII to validate our proposed approach. We use all machines available and divide them into two groups. When the signal of interest is stationary: containing the fan and pump; and when the signal of interest is non-stationary: containing the valve and the slider. For example, in the case of the valve, the acoustic emissions are impulsive and sparse. This distinction is important because we illustrate in the results why the nnCP is particularly adapted for stationary signal, but not necessary for non-stationary ones. 
For each machine, we generated training and validation datasets in the same manner as in the paper \cite{purohit2019mimii}. The training dataset $\mathbfcal{X}^{training} $ comprises only normal sound. The validation dataset $\mathbfcal{X}^{valid} $ contains as many normal recordings as abnormal ones. The objective is to detect only the abnormal sound from the validation dataset after having trained the network with the training dataset.
\subsection{Validation}
The autoencoder presented in Section~\ref{S-det} is trained with the Adam  optimization technique \cite{kingma2014adam} for 50 epochs. We refer to the parameters obtained by solving the problem in Eq.~(\ref{E-cost}) using the $\mathbfcal{X}^{training} $ dataset (baseline case) as $\theta^{*}$ . Similarly, we refer to the parameters obtained by solving the problem in Eq.~(\ref{E-cost}) using the denoised $\hat{\mathbfcal{X}}^{training} $ dataset with  the NMF and the nnCP decomposition respectively as $\theta^{*}_{NMF}$ and $\theta^{*}_{nnCP}$ .

For each recording of the validation dataset $\mathbf{X}^{valid}=\mathbf{X}^{valid}_{::n}$ which can be either normal or abnormal, we compare the outputs of the three setups, referred to as  'Baseline',  'NMF' and  'nnCP',  obtained by applying the decision function in Eq.~\ref{E-des} with the following parameters:
\begin{align}
& \bullet \hspace{0.2cm} {\rm 'Baseline':}  \hspace{0.2cm}  \mathcal{F}(\mathbf{X}^{valid},\theta^{*}) \\
& \bullet \hspace{0.2cm}  {\rm 'NMF':}  \hspace{0.2cm}  \mathcal{F}(\hat{\mathbf{X}}^{valid},\theta^{*}_{NMF}) \\
& \bullet \hspace{0.2cm}   {\rm 'nnCP':}   \hspace{0.2cm} \mathcal{F}(\hat{\mathbf{X}}^{valid},\theta^{*}_{nnCP}) 
\end{align}
Please note that for the case 'NMF' and  'nnCP' the validation dataset $\mathbfcal{X}^{valid} $ is denoised using the NMF or the nnCP decomposition respectively. The number of components $K$ used is the same as for the denoising of the training dataset. NMF and nnCP are applied on all datasets independently. 

The comparison between the decision obtained from all recordings of the validation dataset to the ground truth is made via area under the receiver operating characteristic curve (AUC). Each experiment is repeated 5 times with different weight initialisations for the neural network.

\subsection{Stationary case}
\begin{figure*}
\centering
\includegraphics[width=\textwidth]{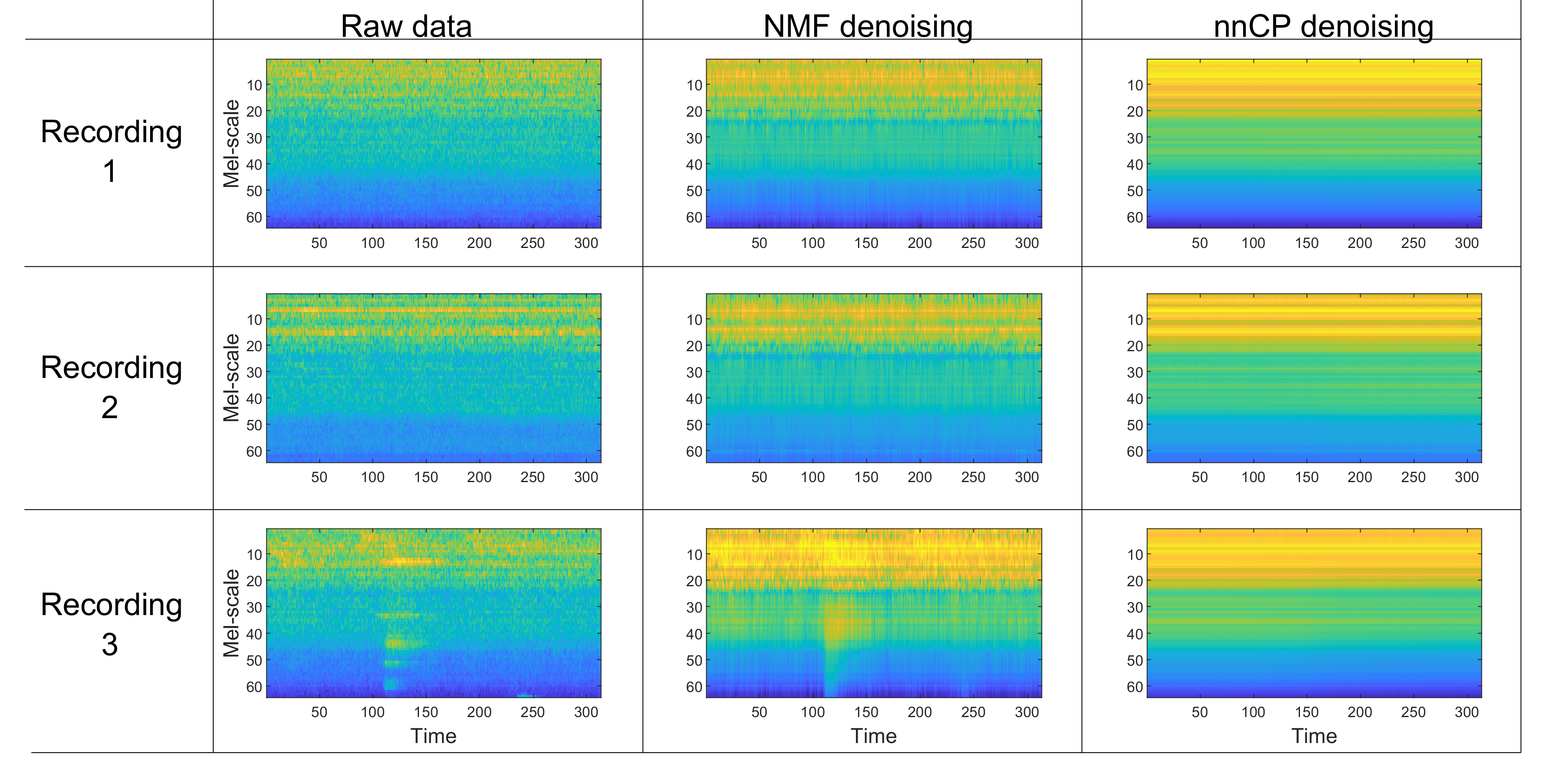}
\caption{Table of figures where rows are three different recordings of the machine fan id 02. The columns show the influence of the denoising strategies with NMF and nnCP case for each recording.}\label{F-sta}
\end{figure*}

We consider machines under stationary operation  in the MIMII dataset: the fan and the pump.  Figure~\ref{F-sta} shows the impact of the denoising for three recordings of the Fan machine id 02. To illustrate the denoising performances, we choose empirically $K=10$ for the NMF and $K=20$ for the nnCP. The first recording doesn't present any particularities, we can see that even with a higher number of components $K$ than for the NMF, the nnCP seems to remove more noise. The second recording contains different spectral motifs compared to the first one. In this case,  the nnCP can retrieve every spectral motif, which is not the case for the NMF (a low-frequency motif is missing). Finally, the third recording presents an artefact delimited in time. This artefact is eliminated with the nnCP. Indeed, every event that is not present in a consequent number of recordings in the same period will be removed (if $K$ is not too high). This demonstrates the advantage of nnCP to denoise stationary processes as it will remove one-off and temporally inconsistent events.

The Table~\ref{F-sta2} shows the mean of the AUC for the different fan and pump datasets available for the three applied approaches 'Baseline', 'NMF' and 'nnCP'. We consider dataset with a medium signal to noise ratio (SNR=0dB) and a low SNR (SNR=-6dB). For the 'NMF' and 'nnCP' approaches, we apply grid search $K=[5,10,20]$ and only retain for each machine the best mean a posteriori based on the anomaly detection performance. We can observe that the best result is often obtained using $K=20$ for the nnCP. More refined results could be found by optimizing $K$ based on a validation dataset. In every case, the nnCP provides a sensible improvement compared the baseline and the NMF method (except for the pump id 06 with SNR=-6dB). We have an average improvement of the AUC between the nnCP approach and the baseline of 0.13 for a SNR of 0 dB and 0,9 for a SNR of -6 dB. These results demonstrate the relevance of the nnCP denoising to lower the complexity of the dataset for stationary machines. 

\begin{figure*}
\centering
\begin{subfigure}[b]{0.32\textwidth}
        \centering \includegraphics[width=\columnwidth]{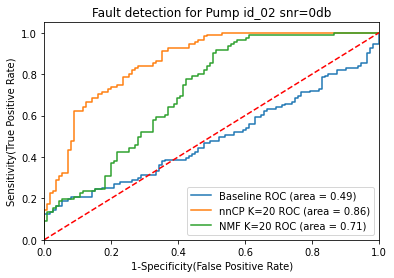}
        \caption{}
   \end{subfigure}
\begin{subfigure}[b]{0.32\textwidth}
        \centering \includegraphics[width=\columnwidth]{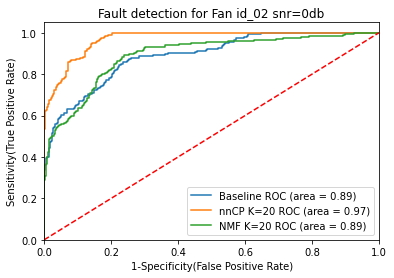}
        \caption{}
   \end{subfigure}
   \begin{subfigure}[b]{0.32\textwidth}
        \centering \includegraphics[width=\columnwidth]{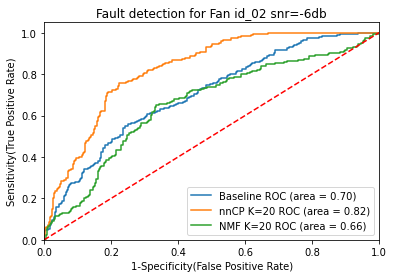}
        \caption{}
   \end{subfigure}

\caption{ROC curves for the 'baseline', 'NMF' and 'nnCP' denoising procedures for (a) the pump id 00 with 0dB (b) the fan id 02 with 0 dB and (c)  the fan id 02 with 6 dB}\label{F-sta3}
\end{figure*}

Figure~\ref{F-sta3} provides three ROC curves obtained the three approaches It illustrates, for these example, that the nnCP denoising decreases at the same time the false-positive rate and the false-negative rate. We obtain similar improvements also for other machines.


\renewcommand{\arraystretch}{1.4}
\begin{table}
\centering
\caption{Mean of AUC for the 5 experiments performed with each stationary machine and with the  'baseline', 'NMF' and 'nnCP' procedures.}\label{F-sta2}
\begin{tabular}{|c|c|c|c|c|}
\hline 
Machine & SNR &  {\color{blue}Baseline} &{\color{c3}NMF}  &{\color{c5}nnCP}    \\
\hline
\hline
\multirow{2}{.2\columnwidth}{\centering Pump-00}
& 0 dB & 0.62 & 0.67  & \textbf{0.86}  \\
\cline{2-5} 
& -6 dB  & 0.59 &           0.69               & \textbf{0.74  }\\ 
\hline
\multirow{2}{.2\columnwidth}{\centering Pump-02}
& 0 dB &  0.49  & 0.72 & \textbf{0.88}  \\
\cline{2-5} 
& -6 dB  & 0.53 &           0.65              & \textbf{0.73  }\\
\hline
\multirow{2}{.2\columnwidth}{\centering Pump-04}
& 0 dB & 0.94  & 0.97 & \textbf{0.98}  \\
\cline{2-5} 
& -6 dB  & 0.91 &           \textbf{0.93 }             & \textbf{0.93  }\\
\hline
\multirow{2}{.2\columnwidth}{\centering Pump-06}
& 0 dB & 0.81 & 0.83 & \textbf{0.87}  \\
\cline{2-5} 
& -6 dB  & \textbf{0.65} &         0.50         & 0.63 \\
\hline
\multirow{2}{.2\columnwidth}{\centering Fan-00}
& 0 dB & 0.63 & 0.57 & \textbf{0.70}  \\
\cline{2-5} 
& -6 dB  &0.56 &         0.51         &\textbf{ 0.61 }\\
\hline
\multirow{2}{.2\columnwidth}{\centering Fan-02}
& 0 dB &0.87 & 0.87 &  \textbf{0.96} \\
\cline{2-5} 
& -6 dB  &0.69 &         0.64         &\textbf{ 0.77 }\\
\hline
\multirow{2}{.2\columnwidth}{\centering Fan-04}
& 0 dB &0.77 & 0.72 &  \textbf{0.90} \\
\cline{2-5} 
& -6 dB  &0.59 &         0.50        &\textbf{ 0.68 }\\
\hline
\multirow{2}{.2\columnwidth}{\centering Fan-06}
& 0 dB & 0.99 &  \textbf{1.00}  &  \textbf{1.00} \\
\cline{2-5} 
& -6 dB  &0.88 &         0.90        &\textbf{ 0.97 }\\
\hline
\hline
\multirow{2}{.2\columnwidth}{\centering \textbf{Average}}
& 0 dB &  0.77 & 0.80 & \textbf{0.90}  \\
\cline{2-5} 
& -6 dB  &0.67 &         0.66        &\textbf{ 0.76 }\\
\hline
\end{tabular} 
\end{table}

\subsection{Non-stationary case}
\begin{figure*}
\centering
\includegraphics[width=\textwidth]{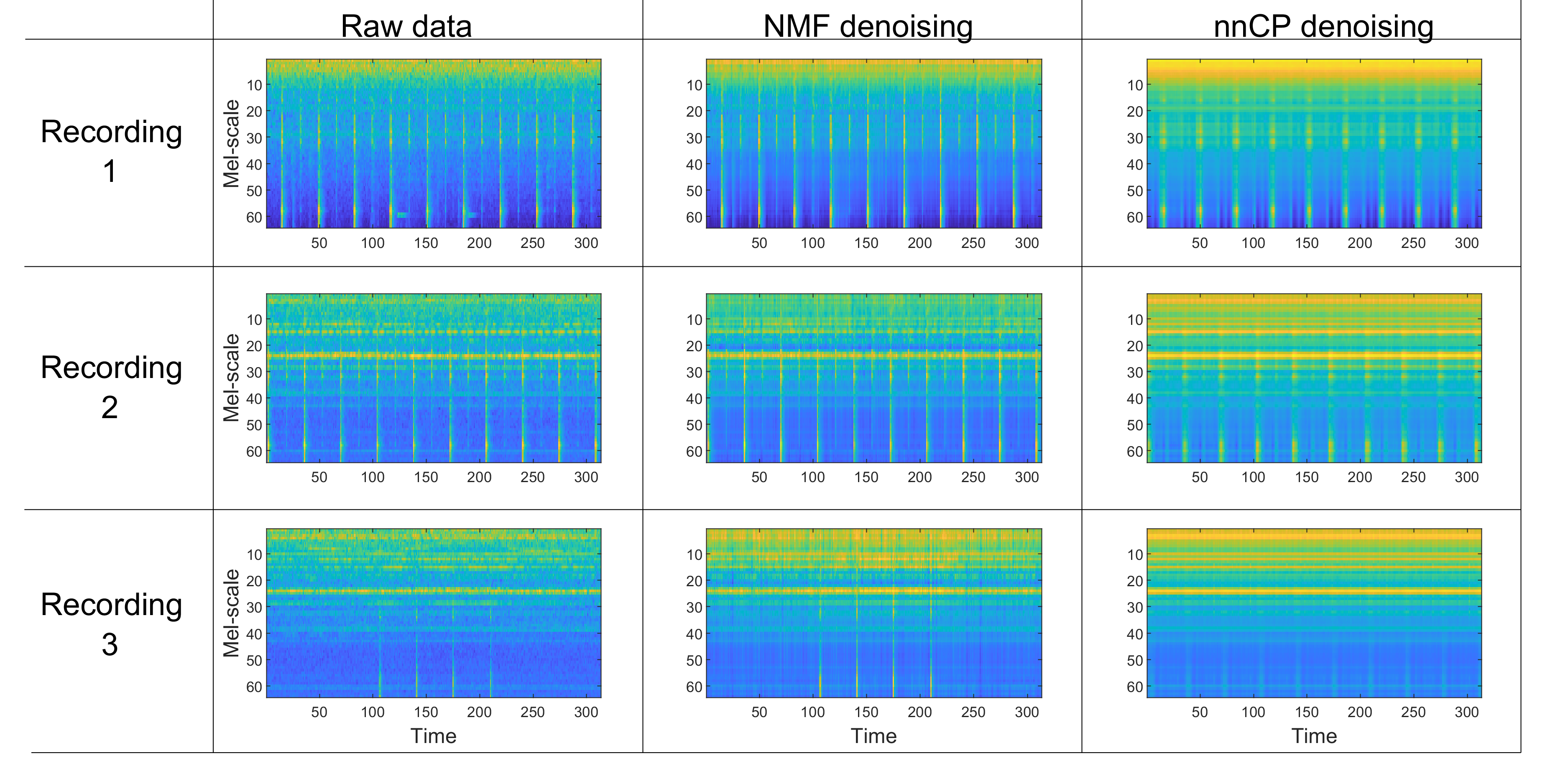}
\caption{Table of figures where rows are three different recordings of the machine valve id 00. The columns show the influence of the denoising strategies with NMF and nnCP for each recording.}\label{F-nons}
\end{figure*}

We consider now the non-stationary machines. We show the impact of the denoising in Figure~\ref{F-nons} for three recordings of the machine valve id 00.  As illustrated by the recording 1 and 2, the valve acoustic emission is characterised by sparse and periodic events. However, from one recording to another, the spike does not activate at the same time periods. Thus, the NMF can retrieve these spikes but not the nnCP. For the nnCP, the spikes are smoothed to account for several recordings that have spike activation approximatively at the same time. However, the quality of spike reconstruction is drastically impaired. For the last recording (corresponding to an abnormal sound), the spike does not activate periodically anymore, they are all removed by the nnCP method because this one-off configuration of the spikes is considered as an artefact. 

The Table~\ref{F-nons2} shows the mean of the AUC for the different valve and slider datasets available for the three approaches 'Baseline', 'NMF' and 'nnCP'. We only consider datasets with medium SNR (SNR=0dB). These results demonstrate that the nnCP decomposition is not adapted for these kinds of non-stationary signals. However, most of the time, the NMF leads to an improvement of the baseline

\begin{table}
\centering
\caption{Mean of AUC for the 5 experiments performed with each non-stationary machine and with the  'baseline', 'NMF' and 'nnCP' procedures.}\label{F-nons2}
\begin{tabular}{|c|c|c|c|}
\hline 
Machine &  {\color{blue}Baseline} &{\color{c3}NMF}  &{\color{c5}nnCP}    \\
\hline
\hline
Valve-00 & 0.62 &\textbf{ 0.69}  & 0.47  \\
\hline
Valve-02 & 0.59  & 0.59 & \textbf{0.62}  \\
\hline
Valve-04 & \textbf{0.65}  & 0.61 &  0.48 \\
\hline
Valve-06 & 0.67 & \textbf{0.76} &   0.56 \\
\hline
Slider-00 & 0.98 &\textbf{ 1.00} & 0.93 \\
\hline
Slider-02 & 0.83 & 0.86 & \textbf{0.90}  \\
\hline
Slider-04 & 0.81 & \textbf{ 0.82} &  0.70 \\
\hline
Slider-06 & 0.55 & \textbf{0.62 }&  0.45 \\
\hline
\hline
Average & 0.71 &\textbf{ 0.74 }& 0.64  \\
\hline
\end{tabular} 
\end{table}

\section{Conclusion and perspectives}\label{S-con}
In this work, we focus on the importance of pre-processing the data before using a deep learning method. We propose two decompositions, the non-negative matrix factorisation (NMF) and the non-negative canonical polyadic decomposition (nnCP) as data-driven, quasi non-parametric modules for spectral data denoising prior to using a neural network. We show on a acoustic emmission monitoring case study, the  MIMII baseline, that nnCP reduces sensibly the complexity of the dataset, and considerably improves the MIMII baseline in the case of machines emitting stationary sound. However, the nnCP seems unsuitable when the sound is non-stationary. 


Two main exciting research paths could be pursued from this work. 

Concerning the limited applicability of the nnCP to non-stationary cases, a first approach would be to centre the acoustic emission of the machine for each recording. However, this pre-processing can be time-consuming. Another approach could be to explore more deeply the use of the NMF decomposition since it showed interesting results compared to the baseline for the non-stationary cases. Finally, an extension of the nnCP for the non-stationary case could be proposed by imposing the time mode of the decomposition to be shift-invariant \cite{tang_sparse_2014}, \cite{kuang2015multi}. 

One of the main limitations of the method is the fact that after the training, a new recording must be denoised before applying the decision rule. Indeed, the network was trained on the denoised dataset, with a limited amount of noise and artefacts. Then, a large number of new recordings have to be regrouped to be effectively denoised with the nnCP decomposition. To overcome this problem, an online version of the nnCP, where its components are updated each time a new recording is considered, could be proposed \cite{mairal2010online}. 



\section*{Acknowledgment}
This study was supported by the Swiss Innovation Agency (lnnosuisse) under grant number: 47231.1 IP-ENG.

\section*{Nomenclature}
In order to facilitate the reading of the article, lower-case letter are used for scalars ($x$), vector are in bold ($\mathbf{x}$), matrix are in bold capital ($\mathbf{X}$) and tensor are in bold calligraphic letter ($\mathbfcal{X}$). $f$, $t$, $n$ and $k$ are indices and $F$, $T$, $N$ and $K$ denote their index upper bounds.

\begin{tabular}{ l  l }
	$\mathbfcal{X}$			& Multi-recording time-frequency representation\\ 
	$\mathbf{X}$			&Time-frequency rep. of one recording\\ 
	$\mathbf{x}$		& Frequencies for one frame \\ 
	$\mathbf{A}$			& Spectral components \\  
	$\mathbf{B}$			& Temporal components\\ 
	$\mathbf{C}$			& Recording components\\ 
	$f$			& frequency index\\ 
	$t$  	   		& time (frame) index \\ 
	$n$	   		& recording index \\ 
	$k$      		& component index\\ 
	$\theta$	   		&Parameters of the neural network\\ 
	    $E(\bullet )$   	   	& Encoder network\\ 
	    $D(\bullet )$   	   	& Decoder network\\ 
	$\mathcal{L}(\bullet )$	   		& Loss function\\ 
	$\mathcal{F}(\bullet )$	   		& Decision function\\ 
 \end{tabular}

\bibliographystyle{apacite}
\PHMbibliography{MyCollection}
\end{document}